\documentclass{article} %
\usepackage[table]{xcolor}

\usepackage{iclr2024_conference,times}

\usepackage{hyperref}
\usepackage{url}

\usepackage[utf8]{inputenc} %
\usepackage[T1]{fontenc}    %
\usepackage{hyperref}       %
\usepackage{url}            %
\usepackage{booktabs}       %
\usepackage{amsfonts}       %
\usepackage{nicefrac}       %
\usepackage{microtype}      %
\usepackage{natbib}
\usepackage{amsmath}
\usepackage{amssymb}
\usepackage{mathtools}
\usepackage{amsthm}
\usepackage{microtype}
\usepackage{graphicx}
\usepackage{subfigure}
\usepackage{booktabs} %
\usepackage{placeins}
\usepackage{multicol,multirow,array,graphicx}
\usepackage{fourier} 
\usepackage{array}
\usepackage{makecell}
\definecolor{lightgray}{gray}{0.9}

\newcommand{\FPR}{FPR}
\newcommand{\FPRQ}{FPR@TPRQ}

\title{How to train your ViT for OOD detection}

\iclrfinalcopy %

\author{%
   Maximilian Müller
   and Matthias Hein\\
   University of Tübingen and Tübingen AI Center \\
   \texttt{maximilian.mueller@wsii.uni-tuebingen.de}
}

\begin{document}

\maketitle

\begin{abstract}
VisionTransformers have been shown to be powerful out-of-distribution detectors for ImageNet-scale settings when finetuned from publicly available checkpoints, often outperforming other model types on popular benchmarks. In this work, we investigate the impact of both the pretraining and finetuning scheme on the performance of ViTs on this task by analyzing a large pool of models. We find that the exact type of pretraining has a strong impact on which method works well and on OOD detection performance in general. We further show that certain training schemes might only be effective for a specific type of out-distribution, but not in general, and identify a best-practice training recipe. 
\end{abstract}

\section{Introduction}
\label{sec:intro}
Deep neural networks have undeniably achieved remarkable success across a spectrum of real-world applications, showcasing outstanding performance. Nevertheless, they often exhibit unforeseen behaviour when confronted with unknown situations like receiving an input that is not related to the task it has been trained on. These anomalous samples are commonly known as out-of-distribution (OOD) data, and deep neural network classifiers often make unwarrantedly confident predictions, incorrectly categorizing them within one of the in-distribution (ID) classes \citep{hendrycks2017MSP, HeiAndBit2019}. 

In response to this challenge, the development of OOD detectors has emerged as a research field. These detectors serve the purpose of identifying OOD inputs, while at the same time letting ID inputs pass through, such that depending on the application appropriate measures can be taken. 
Recent research has shown that VisionTransformers can perform particularily well on this task. An investigation by \cite{anonymous2023COOD} found that "a subset of ViTs achieves the best [...] OOD detection performance" in their benchmark under various training regimes. However, not all training strategies yielded competitive results. Further, OOD detection methods that directly access pre-logit feature information, such as the Mahalanobis distance \cite{LeeMahalanobis2018}, were excluded from their study, presumably due to the increased computational cost arising from the necessity of iterating over the ImageNet \citep{imagenet_cvpr09} train set for feature extraction. However, \cite{bitterwolf2023ninco}, who also identified certain ViTs as the top performers in their benchmark, reported that precisely these feature-based methods exhibit the highest efficacy, particularly when coupled with supervised pretraining on the ImageNet-21k dataset. They further hypothesize that the choice of pretraining might have a profound impact on OOD detection performance.

Motivated by these findings, the primary objective of this paper is to undertake an extensive exploration of the role played by pretraining and, more broadly, training hyperparameters for ViTs in the context of OOD detection.

\begin{figure}
\vspace{-0.3cm}
    \centering
    \includegraphics[width=1.\textwidth]{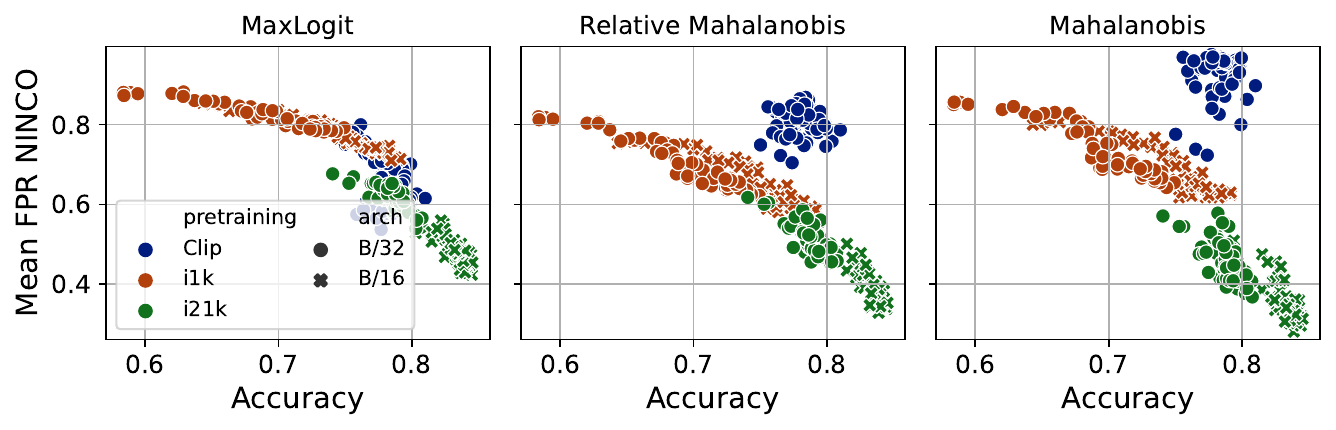}
        \includegraphics[width=1.\textwidth]{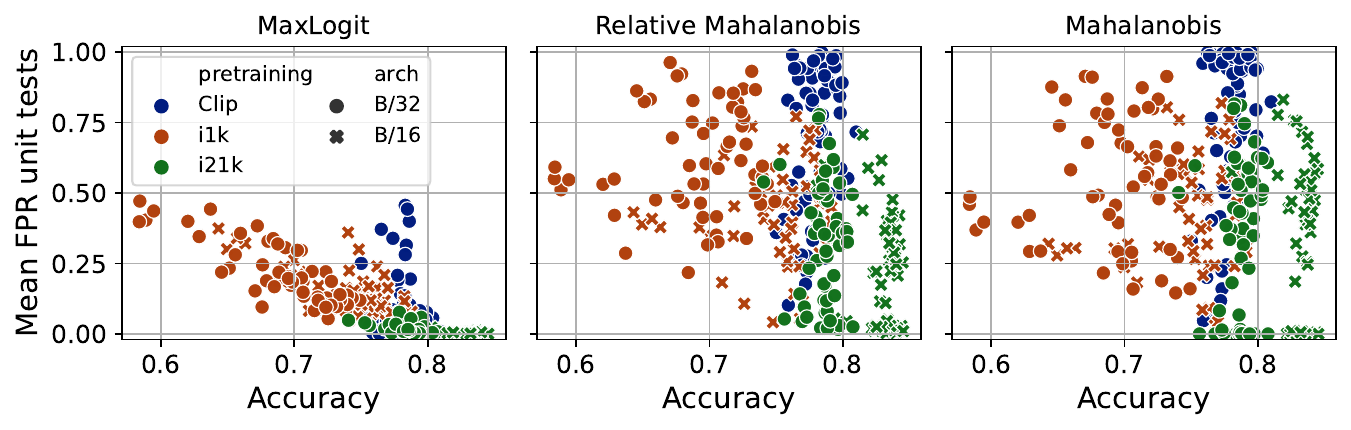}
    \caption{\textbf{Pretraining matters:} 
    ImageNet-21k pretraining paired with Mahalanobis-based detection methods strongly outperforms other detectors on NINCO, yet fails in many cases on the unit-test task.}
    \label{fig:NINCO-all-pre}
    \vspace{-0.25cm}
\end{figure}

\section{Experimental Setup}
In order to investigate the role of pretraining and hyperparameters on large-scale OOD detection in a controlled setting, we combine two publicly available model pools 
and evaluate them with commonly used post-hoc OOD detection methods. We report FPR values in the main paper, but AUC values can be found in the Appendix and lead to the same conclusions.

\textbf{Models. } We evaluate ViT-B/16-224 and ViT-B/32-224 weights from \cite{steinerhowtotrainyourvit}, which were trained from scratch on ImageNet-1k or ImageNet-21k with varying hyperparameter settings (including augmentations, weight decay, and stochastic depth / dropout) in a supervised fashion, and subsequently finetuned on ImageNet-1k for 20.000 steps with 2 different learning rates, leading to a rich pool of 222 models.
We further evaluate models from \cite{modelsoupswortsman22a}, who investigated the benefits of "souping" models that were finetuned from a common checkpoint with different hyperparameters. To this end, they finetuned a CLIP-pretrained ViT-B/32-224 on ImageNet-1k with different settings (varying augmentations, mixup, weight decay, lr, finetuning epochs and smoothing). \citet{modelsoupswortsman22a} eventually combine them either in a uniform or greedy fashion for enhanced performance, and we evaluate both the individual and the souped models. 
Overall, our pool comprises 296 different models (details in Appendix~\ref{app:exp-details}).

\textbf{Dataset.}
In a recent study \cite{bitterwolf2023ninco} showed that most of the commonly used datasets for OOD detection evaluation for ImageNet-1k are severely flawed and contain a significant fraction of ID samples. They proposed the hand-cleaned \textit{NINCO} dataset as a solution, which we will therefore use for our evaluations. \textit{NINCO} consists of real-world images, with several 'hard' (i.e. similar to ImageNet-1k) classes, hence mimicking an OOD test distribution which is in parts very close to the in-distribution. Further, they provide a range of synthetic \textit{unit-test} sets, mimicking noise distributions that might occur in the real world, but should be 'far' from the ImageNet test set. A good detector should be able to solve both tasks, and we will therefore contrast the results on \textit{NINCO} with those on the \textit{unit-tests}.

\textbf{Methods.}
Post-hoc OOD detection methods can be separated into methods that directly access pre-logit feature information (called \textbf{feature-based}) and methods that only leverage logit or softmax information (called \textbf{logit-based}). In the main paper, we show results for the respective best methods: Max-Logit \citep{hendrycks22Scaling}, Mahalanobis \citep{LeeMahalanobis2018} and relative Mahalanobis distance \citep{RenRelMaha2021}. We provide results for more methods in Appendix~\ref{app:more-methods} and give an overview over all methods in Appendix ~\ref{sec:methods}.

\vspace{-0.2cm}
\section{Observations}
\vspace{-0.2cm}
\textbf{Pretraining Dataset. }
In Figure~\ref{fig:NINCO-all-pre} we plot the mean FPR of all investigated models against their in-distribution accuracy for three methods and the two OOD test sets. For Max-Logit, the FPR on NINCO (first row) correlates fairly well with accuracy, irrespective of the pretraining dataset. For the feature-dependent Mahalanobis-based methods, the models with CLIP-pretraining perform significantly worse than models with ImageNet-21k or ImageNet-1k pretraining, even for similar ID accuracy. Those feature-based methods, however, lead to the lowest overall FPR values on NINCO observed in our study when combined with ImageNet-21k pretraining. When considering the supposedly easy unit-test dataset (second row in Figure~\ref{fig:NINCO-all-pre}), the picture changes drastically: Especially for Mahalanobis-based methods the correlation between accuracy and FPR vanishes, and many models with similar test accuracy show vastly different FPR values. Some of the strongest models in terms of accuracy show very poor OOD detection capabilities, often with FPR values over 50\%.
This is concerning, since it indicates that supposedly strong OOD detectors, i.e. ones that are capable of solving the difficult task of separating near-OOD data from ID data, are vulnerable to OOD data from very different distributions. Overall, the strongest OOD detectors are clearly obtained from ImageNet-21k pretraining both for NINCO and the unit-tests, and in many cases CLIP-pretraining can at most slightly improve over ImageNet-1k-only training. In the following, we will go beyond the effects of the pretraining dataset and investigate the role of specific training parameters for each pretraining dataset. 

\begin{figure}
\vspace{-0.4cm}
    \centering
    \includegraphics[width=1.\textwidth]{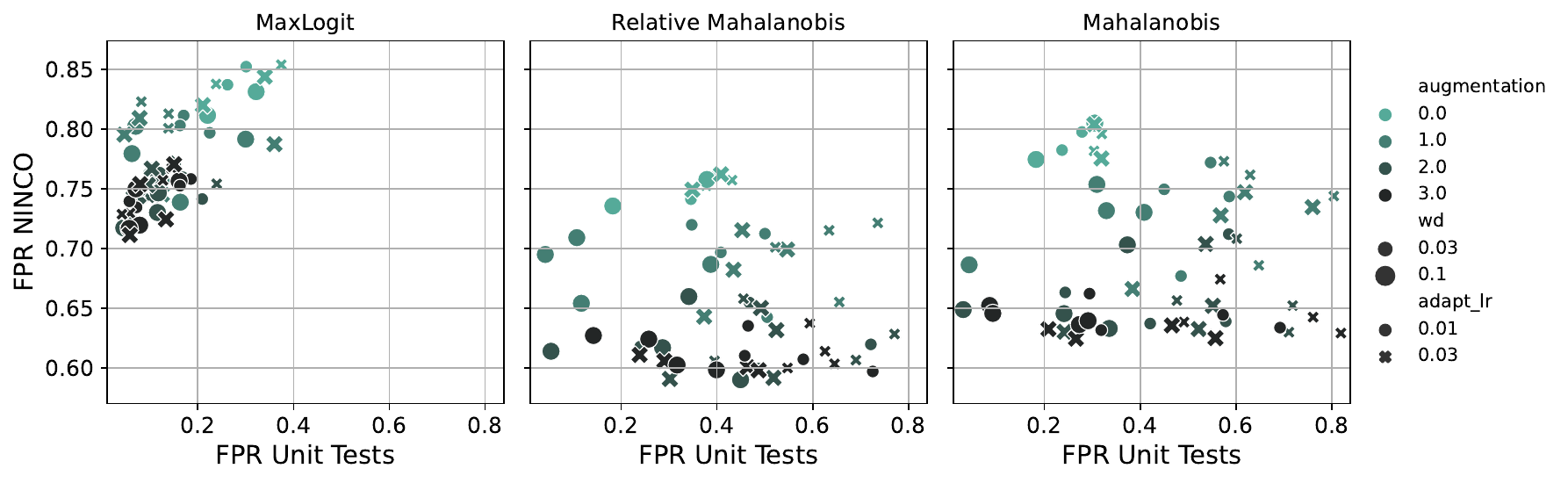}
        \vspace{-0.7cm}
    \caption{ViT-B/16 trained exclusively on ImageNet-1k}
            \label{fig:steiner-vits-1k}
    \vspace{0.3cm}
    \includegraphics[width=1.\textwidth]{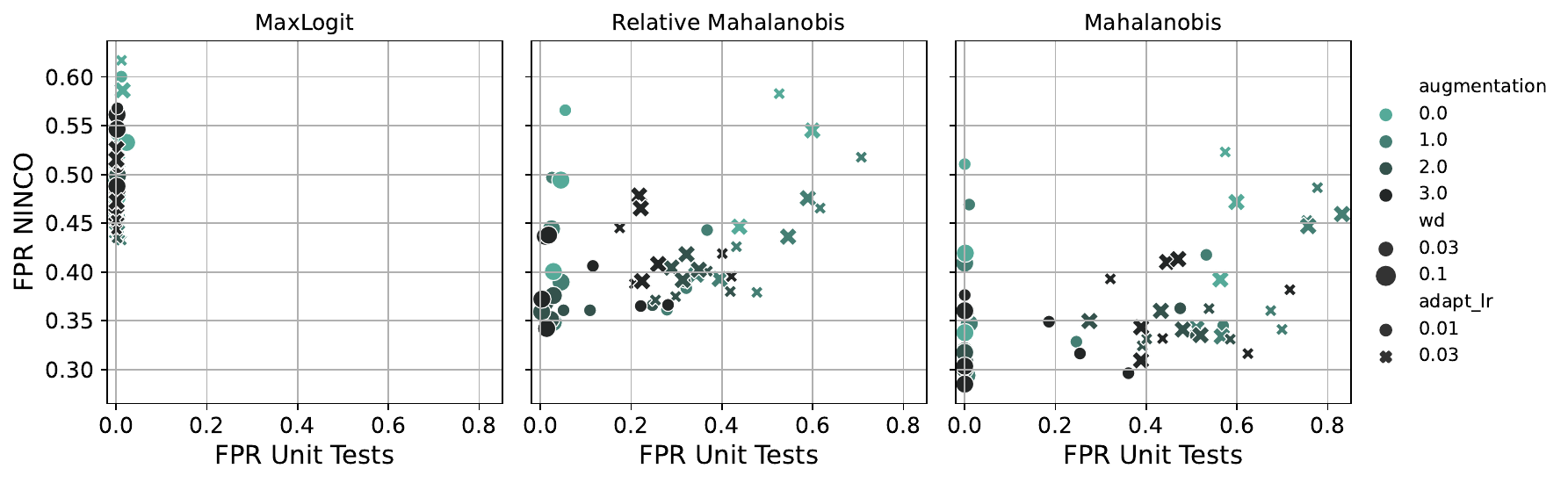}
        \vspace{-0.7cm}
    \caption{ViT-B/16 with pretraining on ImageNet-21k}
    \label{fig:steiner-vits-21k}
    \vspace{-0.25cm}
\end{figure}

\textbf{Pretraining on ImageNet-21k and finetuning on ImageNet-1k. }
When pairing models with ImageNet-21k pretraining (Figure~\ref{fig:steiner-vits-21k}) with logit-based methods like MaxLogit, all models yield FPR values close to 0 for the unit-tests, basically solving the task. There is however a difference on the NINCO dataset, where the FPR values range roughly between 45\% and 60\%. The better models are usually finetuned with larger learning rate, and were pretrained without dropout and with smaller weight decay. Augmentations seem to play less of a role. While the worst models in terms of FPR also show worse accuracy, there are also high-accuracy models with high FPR. Those were typically trained with strong augmentations and dropout and finetuned with small learning rate.

When pairing the models with Mahalanobis distance as OOD detection method, the overall best results on the NINCO dataset are achieved, with the best FPR values below 30\%. 
There are however strong differences between the specific detectors when considering the unit tests. While models that were pretrained with larger weight decay and finetuned with smaller learning rate lead to detectors that perform well on both NINCO and the unit-tests, models with smaller weight decay often and with larger learning rate always fail on the unit tests, with FPR values between 20\% and 80\%. Concerningly, those models showed some of the highest test accuracies (>84.0\%) in our study. On the contrary, there are models with relatively low test accuracy (e.g. 79\%), but close-to-zero FPR on the unit tests. 
When conditioning on larger weight decay and smaller learning rate, the FPR on NINCO correlates well with pretraining validation accuracy ($\tau=-0.72$), which provides a best-practice training strategy: Selecting models that were pretrained on ImageNet-21k with larger weight decay, and amongst those the ones with highest pretraining validation accuracy, and then finetune them with small learning rate (more in App.~\ref{app:details-21k}).

\textbf{Training exclusively on ImageNet-1k. }
When pairing models trained only on ImageNet-1k (Figure~\ref{fig:steiner-vits-1k}) with MaxLogit, there is a strong correlation between the performance on NINCO and the test accuracy ($\tau=0.88$), i.e. settings that improve the in-distribution accuracy (e.g. stronger augmentations, dropout and to a smaller extent also stronger weight decay) also tend to improve the OOD detection capabilities for NINCO. While the correlation for the unit tests is smaller, stronger augmentations still lead
to better performance. 

When pairing models trained only on ImageNet-1k with Mahalanobis, lower FPR values on NINCO can be achieved compared to logit-based methods. Stronger augmentations lead to stronger performance on NINCO (and also higher test accuracy). However, there are many cases where the models completely fail on the unit tests (FPR values >65\%), even when pretrained with strong augmentations. Some of those are amongst the best-performing models in terms of test accuracy (when pretrained with dropout, small weight decay and finetuned with large learning rate). The best detectors on the unit tests were pretrained with larger weight decay and finetuned with smaller learning rate, and there are only relatively few models that perfrom well on both NINCO and the unit-tests. For ImageNet-1k pretraining the \textit{Relative} Mahalanobis distance works slightly better than the conventional Mahalanobis distance, but the overall trends remain similar.

\textbf{Pretraining on CLIP and finetuning on ImageNet-1k. }
For models pretrained with CLIP (Figure~\ref{fig:clip} in the appendix), feature-based methods like Mahalanobis distance (but also other methods like ViM as shown in appendix \ref{app:more-methods}) bring no improvements over logit-based methods, with FPR values commonly above 90\%. The best method for those models is MaxLogit, but other logit-based methods (e.g. Energy, KL-Matching, ...) perform similar. In general, accuracy does not correlate well with FPR, and often there is no significant difference in the OOD detection capabilites of models with fairly different test accuracies. The souped models bring no consistent advantage over the individual models. 
For MaxLogit, many models achieve 0\% FPR on the unit tests, and all models with strong rand-aug perform relatively well (even though also models without strong rand-aug \textit{can} achieve 0\% FPR). 

\textbf{Do we need finetuning at all?} Due to the strong performance of ViTs with ImageNet-21k pretraining it is natural to ask whether training \textit{only} on ImageNet-21k might be enough to obtain strong feature-based detectors for ImageNet-1k. In Table~\ref{tab:big-table-vits} in App.~\ref{app:details-21k} we report the performance of the 21k-checkpoints with Mahalanobis distance \textit{before} finetuning and show that this is not the case: Finetuning improves the detection performance consistently. We provide an extended discussion on this in App.~\ref{app:details-21k}.
\vspace{-0.15cm}
\section{Conclusions}
\vspace{-0.15cm}
In this study we showed that for ViTs the effectiveness of an OOD detection method depends crucially on the detailed pretraining and finetuning scheme of the model. While pretraining on ImageNet-21k yields clearly the best detectors (as also observed for generalization tasks by \cite{steinerhowtotrainyourvit}), many detectors, especially feature-based ones, are susceptible to small changes in the hyperparameters and might only work for a particular type of out-distribution.

In their work, \citet{steinerhowtotrainyourvit} recommended choosing the ImageNet-21k pretrained checkpoints with highest upstream validation accuracy for finetuning on downstream tasks. We show that this is not enough for OOD detection, since e.g. a too large learning rate can result in poorly performing Mahalanobis-based detectors, even though the validation accuracy both before and after finetuning might be comparably high (more details can be found in App~\ref{app:details-21k}).

Consistently good results were obtained for Mahalanobis distance as OOD detection method with models pretrained on ImageNet-21k with larger weight decay during pretraining and 
subsequent finetuning with small learning rate. We hypothesize that larger weight decay might help to structure the feature space in a beneficial way during supervised pretraining, and that a smaller finetuning learning rate might preserve this structure better. CLIP pretraining, which is based on a cosine-similarity measure, seems to impose a different structure, which does not lend itself towards the current feature-based OOD detection methods.

\subsubsection*{Acknowledgments}
We acknowledge support from the German Federal Ministry of Education and Research (BMBF) through the Tübingen AI Center (FKZ: 01IS18039A) and from the Deutsche Forschungsgemeinschaft (DFG, German Research Foundation) under Germany’s Excellence Strategy (EXC number 2064/1, Project number 390727645), as well as from the Carl Zeiss Foundation in the project “Certification and Foundations of Safe Machine Learning Systems in Healthcare” and the European Laboratory for Learning and Intelligent Systems (ELLIS).

\bibliography{bib.bib}
\clearpage

\section{Supplementary Material}

\appendix
\section{Experimental details}\label{app:exp-details}
We report the detailed hyperparameter grid for the models from \citet{steinerhowtotrainyourvit} in Table~\ref{tab:params-steiner} and from \citet{modelsoupswortsman22a} in Table~\ref{tab:params-soup}. For the plots in the main paper except for Figure~\ref{fig:NINCO-all-pre}, we show only the B/16 variants of the \citet{steinerhowtotrainyourvit} models for better visibility, but the observed trends remain the same for the B/32 variants. We further aggregate the augmentations from Table~\ref{tab:params-steiner} into \{none,light, medium, strong\}. For CLIP-models (e.g. in Figure~\ref{fig:clip}), we plot the \textit{magnitude} of rand-aug, assume min-aug to have magnitude 0, and discard the number of augmentations. We further aggregate mixup into weak ($\leq0.15$) and strong ($>0.15$) for easier visualization.
\FloatBarrier

\begin{table}[h]
    \centering
        \caption{Hyperparameter grid for the selected ViTs from \citet{steinerhowtotrainyourvit}. Hyperparameters vary for both pretraining and finetuning.}

    \begin{tabular}{c|c|c|c|c|c|c|c|c|c}
    &\multicolumn{6}{c|}{\textbf{Pretraining}} & \multicolumn{3}{c}{\textbf{Finetuning}}\\
    \hline
          arch & epochs & lr & dset & wd & aug & \makecell{dropout / \\stoch. d.} & lr & steps & res.   \\
         \hline
          \makecell{B/16,\\B/32} & 300 & 0.001 & \makecell{IN-1k, \\IN-21k} & \makecell{0.03,\\0.1} & \makecell{strong1,strong2 \\med1,med2 \\light0,light1\\ none} & \makecell{0,\\0.1} & \makecell{0.01,\\0.03} & 20k & 224 
    \end{tabular}
    \label{tab:params-steiner}
\end{table}

\begin{table}[h]
    \centering
    \caption{Hyperparameter grid for the CLIP-pretrained ViTs from \citet{modelsoupswortsman22a}. Hyperparameters vary only for finetuning.}
    \begin{tabular}{c|c|c|c|c|c|c|c|c}
    &\multicolumn{8}{c}{\textbf{Finetuning}}\\ \hline
    arch &aug & \makecell{r-aug\\n} & \makecell{r-aug\\m} & mixup & smoothing & epochs & wd & lr \\ \hline
    B/16 & \makecell{min with  \\
    prob 1/3\\ else r-aug} & 0,1,2 & $[0,20]$ & \makecell{0 with \\prob 0.5,\\else\\
    $[0,0.9]$} & \makecell{0 with \\prob 0.5,\\else\\
    $[0,0.25]$} & [4,16] & \makecell{$10^{-\lambda},$ \\ $\lambda\in[4,6]$} & \makecell{$10^{-\lambda},$ \\$\lambda\in[0.2,4]$} \\
    \end{tabular}
    \label{tab:params-soup}
\end{table}

\FloatBarrier
\section{Additional Results}
\subsection{Pretraining plots with AUC}\label{app:auc-pretraining}
\begin{figure}
\vspace{-0.2cm}
    \centering
    \includegraphics[width=1.\textwidth]{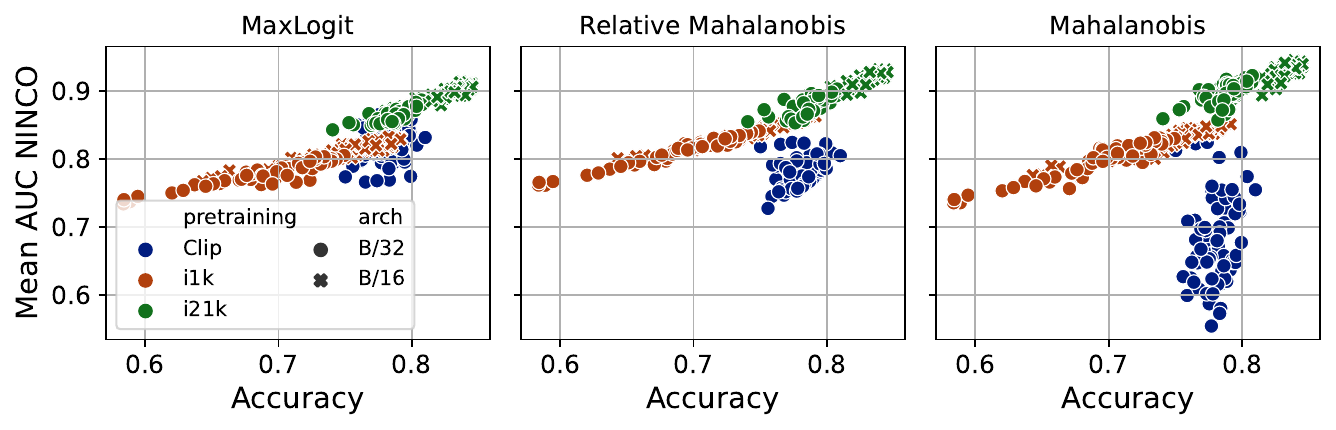}
        \includegraphics[width=1.\textwidth]{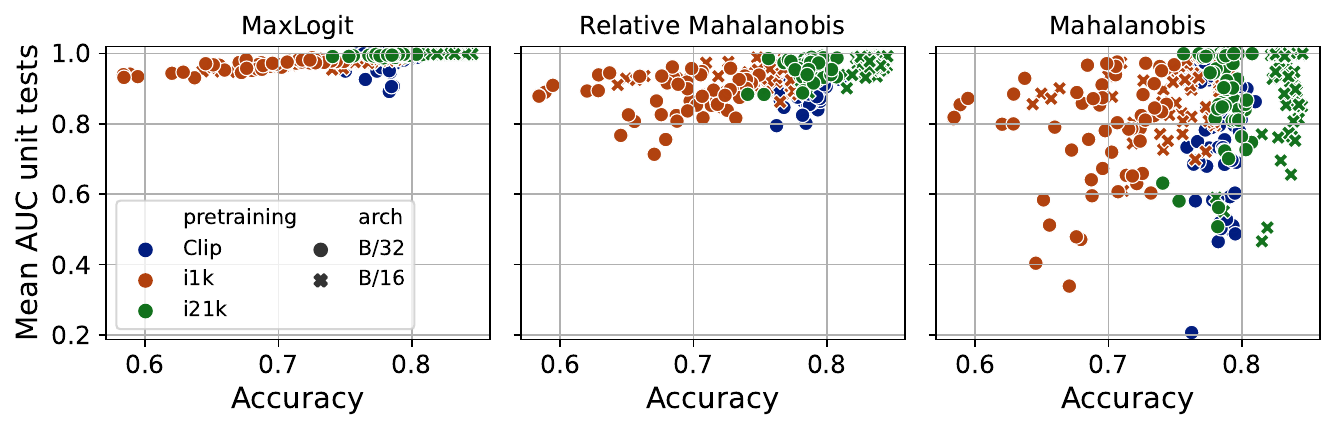}
    \caption{\textbf{Pretraining matters:} 
    ImageNet-21k pretraining paired with Mahalanobis-based detection methods strongly outperforms other detectors on NINCO, yet fails in many cases on the unit-test task.}
    \label{fig:NINCO-all-pre-auc}
    \vspace{-0.2cm}
\end{figure}

In Figure \ref{fig:NINCO-all-pre-auc} we show the mean AUC on NINCO and the unit tests vs. in-distribution accuracy to complement the FPR plot from the main paper. The conlusions remain the same: The best detectors on NINCO are obtained with Mahalanobis distance and ImageNet-21k pretraining, yet some of those fail completely on the unit tests.
\subsection{Detailed results for ViT-B/16 with ImageNet-21k pretraining}\label{app:details-21k}
We report detailed results for the ViT-B/16 models with Imagenet-21k pretraining in Table~\ref{tab:big-table-vits} (FPR) and Table\ref{tab:big-table-vits-auroc} (AUC), since this setup yields the overall strongest detectors. The Table is sorted according to the validation accuracy of the pretraining checkpoints (higher on top), since \citet{steinerhowtotrainyourvit} suggest selecting the models with highest pretraining accuracy for finetuning on downstream tasks. While this works relatively well for OOD detection on NINCO (the top models all achieve FPR values around 30\% with Mahalanobis distance), the effect of the learning rate on Mahalanobis distance is striking: For most checkpoints, finetuning with a learning rate of 0.01 (white rows) leads to FPR values of around 0\% (solving the unit-test task), whereas finetuning with a learning rate of 0.03 (grey rows) never yields FPR values below 30\%, and often even much larger ones, outlining that the structure of the feature space that lends itself towards Mahalanobis distance seems to be crucially affected by the learning rate, even though this might not strongly affect the models generalization performance. 

In order to achieve good performance on both NINCO and the unit tests, the best strategy is thus to select models that were pretrained on ImageNet-21k with larger weight decay, and amongst those the ones with highest pretraining validation accuracy, and then finetune them with small learning rate and combine them with Mahalanobis distance as detection method. One remaining question is how to achieve strong pretraining accuracy. This is thoroughly discussed in \citet{steinerhowtotrainyourvit} and depends non-trivially on the interplay between augmentation, regularization and model capacity. For our study, we observed that good results could be obtained, as long as strong augmentations are not paired with dropout, and none or light augmentations are not paired with zero dropout. Arguably, this might however change when considering different model sizes. 

Since the most powerful detectors in our survey are obtained with pretraining on ImageNet-21k, the question arises whether training \textit{only} on ImageNet-21k might be enough to obtain strong detectors for ImageNet-1k. We therefore also report the performance of the pretraining checkpoints with Mahalanobis distance, dubbed \textit{Mahalanobis (pre)} in the table. This is, we evaluate the checkpoint \textit{before} finetuning on ImageNet-1k. Due to the different output dimension of ImageNet-1k and ImageNet-21k models (1k vs. 21k classes), evaluating 21k-checkpoints is only possible for methods that operate exclusively in the feature space (like [relative] Mahalanobis distance, knn and cosine). Notably, the performance of Mahalanobis distance after finetuning is \textit{always} better for the respective checkpoints, indicating that the structure of the feature space adapts significantly during the finetuning procedure. Nevertheless, certain pretraining checkpoints yield remarkably good results on NINCO, with FPR values of roughly 40\% - results that are not achieved by any method with CLIP pretraining or ImageNet-1k-exclusive training. On the unit tests, however, the pretraining checkpoints fail consistently, showing FPR values of 40-90\%.

\definecolor{lightgray}{gray}{0.9}
\begin{table}[]
\setlength{\tabcolsep}{3pt}
    \centering
        \caption{Detailed hyperparameters and FPR values for the ViT-B/16 models with ImageNet-21k pretraining. In the \textit{Mahalanobis (pre)} columns we evaluate the pretraining checkpoint (i.e. \textit{before} finetuning) with Mahalanobis distance. As suggested by \cite{bitterwolf2023ninco}, we also report the number of failed unit tests (with FPR values $\geq10\%$).}
    \rowcolors{2}{white}{lightgray}
\resizebox{1.\textwidth}{!}{%
\begin{tabular}{ccccccccccccccc}

 &  &  &  &  &  & \multicolumn{3}{c}{\textbf{MaxLogit}} & \multicolumn{3}{c}{\textbf{Mahalanobis}} & \multicolumn{3}{c}{\textbf{Mahalanobis (pre)}} \\ 
  &  &  &  &  &  & NINCO & \multicolumn{2}{c}{unit tests} & NINCO & \multicolumn{2}{c}{unit tests} & NINCO & \multicolumn{2}{c}{unit tests} \\
  
wd & aug & do & adapt-lr & pre-acc & ft-acc & \multicolumn{1}{c}{{FPR [\%]}} & \multicolumn{1}{c}{{FPR [\%]}} & n failed & \multicolumn{1}{c}{{FPR [\%]}}& \multicolumn{1}{c}{{FPR [\%]}} & n failed & \multicolumn{1}{c}{{FPR [\%]}} & \multicolumn{1}{c}{{FPR [\%]}} & n failed  \\ 
  \hline 
0.1 & medium1 & 0.0 & 0.03 & 50.6 & 83.9 & 46.7 & 0.1 & 0.0 & 34.1 & 48.1 & 14.0 & 41.9 & 71.5 & 14.0 \\ 
 
0.1 & medium1 & 0.0 & 0.01 & 50.6 & 84.5 & 46.5 & 0.0 & 0.0 & 31.9 & 0.0 & 0.0 & 41.9 & 71.5 & 14.0 \\ 
 
0.1 & light1 & 0.0 & 0.03 & 50.4 & 83.5 & 45.4 & 0.1 & 0.0 & 33.4 & 56.7 & 16.0 & 38.6 & 77.2 & 16.0 \\ 
 
0.1 & light1 & 0.0 & 0.01 & 50.4 & 84.4 & 45.7 & 0.0 & 0.0 & 29.4 & 0.5 & 0.0 & 38.6 & 77.2 & 16.0 \\ 
 
0.1 & medium2 & 0.0 & 0.03 & 50.4 & 83.9 & 46.0 & 0.1 & 0.0 & 33.5 & 52.0 & 14.0 & 46.0 & 81.5 & 17.0 \\ 
 
0.1 & medium2 & 0.0 & 0.01 & 50.4 & 84.4 & 48.3 & 0.1 & 0.0 & 30.1 & 0.0 & 0.0 & 46.0 & 81.5 & 17.0 \\ 
 
0.03 & medium2 & 0.0 & 0.03 & 50.4 & 84.5 & 43.5 & 0.2 & 0.0 & 33.1 & 58.5 & 13.0 & 67.9 & 92.9 & 17.0 \\ 
 
0.03 & medium2 & 0.0 & 0.01 & 50.4 & 84.0 & 49.7 & 0.6 & 0.0 & 33.5 & 50.9 & 12.0 & 67.9 & 92.9 & 17.0 \\ 
 
0.03 & strong1 & 0.0 & 0.03 & 50.3 & 84.2 & 44.4 & 0.1 & 0.0 & 31.6 & 62.4 & 14.0 & 68.3 & 95.4 & 17.0 \\ 
 
0.03 & strong1 & 0.0 & 0.01 & 50.3 & 84.0 & 48.5 & 0.1 & 0.0 & 29.6 & 36.1 & 11.0 & 68.3 & 95.4 & 17.0 \\ 
 
0.03 & strong2 & 0.0 & 0.03 & 50.2 & 84.1 & 45.3 & 0.0 & 0.0 & 33.2 & 43.6 & 11.0 & 74.6 & 95.5 & 17.0 \\ 
 
0.03 & strong2 & 0.0 & 0.01 & 50.2 & 83.7 & 51.4 & 0.0 & 0.0 & 31.6 & 25.4 & 7.0 & 74.6 & 95.5 & 17.0 \\ 
 
0.1 & strong1 & 0.0 & 0.03 & 49.9 & 83.5 & 45.9 & 0.0 & 0.0 & 34.3 & 38.9 & 9.0 & 41.7 & 72.9 & 14.0 \\ 
 
0.1 & strong1 & 0.0 & 0.01 & 49.9 & 84.0 & 48.5 & 0.2 & 0.0 & 30.3 & 0.0 & 0.0 & 41.7 & 72.9 & 14.0 \\ 
 
0.03 & medium1 & 0.0 & 0.03 & 49.7 & 84.0 & 43.6 & 0.1 & 0.0 & 36.3 & 53.8 & 14.0 & 58.9 & 94.5 & 17.0 \\ 
 
0.03 & medium1 & 0.0 & 0.01 & 49.7 & 84.3 & 48.9 & 0.0 & 0.0 & 36.3 & 47.5 & 13.0 & 58.9 & 94.5 & 17.0 \\ 
 
0.1 & strong2 & 0.0 & 0.03 & 49.6 & 83.6 & 47.2 & 0.1 & 0.0 & 30.9 & 38.9 & 10.0 & 47.4 & 82.6 & 17.0 \\ 
 
0.1 & strong2 & 0.0 & 0.01 & 49.6 & 83.9 & 48.8 & 0.2 & 0.0 & 28.5 & 0.0 & 0.0 & 47.4 & 82.6 & 17.0 \\ 
 
0.03 & light1 & 0.1 & 0.03 & 49.5 & 83.7 & 44.4 & 0.5 & 0.0 & 34.1 & 69.9 & 15.0 & 77.7 & 97.6 & 17.0 \\ 
 
0.03 & light1 & 0.1 & 0.01 & 49.5 & 83.7 & 46.2 & 0.1 & 0.0 & 32.9 & 24.6 & 6.0 & 77.7 & 97.6 & 17.0 \\ 
 
0.1 & light1 & 0.1 & 0.03 & 49.3 & 83.6 & 44.1 & 0.2 & 0.0 & 34.2 & 51.1 & 12.0 & 53.7 & 72.2 & 17.0 \\ 
 
0.1 & light1 & 0.1 & 0.01 & 49.3 & 84.1 & 45.6 & 0.1 & 0.0 & 30.6 & 0.0 & 0.0 & 53.7 & 72.2 & 17.0 \\ 
 
0.03 & medium1 & 0.1 & 0.03 & 49.2 & 83.9 & 45.5 & 0.1 & 0.0 & 32.4 & 39.1 & 9.0 & 73.6 & 91.5 & 17.0 \\ 
 
0.03 & medium1 & 0.1 & 0.01 & 49.2 & 83.4 & 50.4 & 0.1 & 0.0 & 31.4 & 0.6 & 0.0 & 73.6 & 91.5 & 17.0 \\ 
 
0.1 & light0 & 0.1 & 0.03 & 49.0 & 82.7 & 50.4 & 0.0 & 0.0 & 44.7 & 75.7 & 17.0 & 51.4 & 59.2 & 16.0 \\ 
 
0.1 & light0 & 0.1 & 0.01 & 49.0 & 83.3 & 49.3 & 0.0 & 0.0 & 34.7 & 1.0 & 1.0 & 51.4 & 59.2 & 16.0 \\ 
 
0.03 & light0 & 0.1 & 0.03 & 48.9 & 82.9 & 50.9 & 0.0 & 0.0 & 45.3 & 75.4 & 16.0 & 64.7 & 96.1 & 17.0 \\ 
 
0.03 & light0 & 0.1 & 0.01 & 48.9 & 83.2 & 52.1 & 0.0 & 0.0 & 41.8 & 53.2 & 12.0 & 64.7 & 96.1 & 17.0 \\ 
 
0.03 & light1 & 0.0 & 0.03 & 48.5 & 83.7 & 43.3 & 1.2 & 0.0 & 36.0 & 67.4 & 15.0 & 57.1 & 92.2 & 17.0 \\ 
 
0.03 & light1 & 0.0 & 0.01 & 48.5 & 84.2 & 45.3 & 0.2 & 0.0 & 34.5 & 57.0 & 13.0 & 57.1 & 92.2 & 17.0 \\ 
 
0.03 & medium2 & 0.1 & 0.03 & 48.5 & 83.8 & 46.4 & 0.1 & 0.0 & 33.1 & 40.1 & 10.0 & 84.4 & 96.6 & 17.0 \\ 
 
0.03 & medium2 & 0.1 & 0.01 & 48.5 & 83.2 & 49.5 & 0.1 & 0.0 & 30.9 & 0.6 & 0.0 & 84.4 & 96.6 & 17.0 \\ 
 
0.1 & medium1 & 0.1 & 0.03 & 48.2 & 83.5 & 48.2 & 0.2 & 0.0 & 36.0 & 43.2 & 10.0 & 56.9 & 76.8 & 15.0 \\ 
 
0.1 & medium1 & 0.1 & 0.01 & 48.2 & 83.6 & 48.9 & 0.3 & 0.0 & 31.6 & 0.0 & 0.0 & 56.9 & 76.8 & 15.0 \\ 
 
0.03 & strong1 & 0.1 & 0.03 & 48.1 & 83.3 & 47.9 & 0.1 & 0.0 & 38.2 & 71.7 & 15.0 & 86.7 & 98.5 & 17.0 \\ 
 
0.03 & strong1 & 0.1 & 0.01 & 48.1 & 82.8 & 50.8 & 0.4 & 0.0 & 34.9 & 18.6 & 4.0 & 86.7 & 98.5 & 17.0 \\ 
 
0.1 & medium2 & 0.1 & 0.03 & 47.7 & 83.3 & 47.1 & 0.1 & 0.0 & 35.0 & 27.5 & 6.0 & 62.8 & 88.9 & 17.0 \\ 
 
0.1 & medium2 & 0.1 & 0.01 & 47.7 & 83.3 & 49.9 & 0.3 & 0.0 & 31.8 & 0.0 & 0.0 & 62.8 & 88.9 & 17.0 \\ 
 
0.03 & strong2 & 0.1 & 0.03 & 47.6 & 82.9 & 51.1 & 0.4 & 0.0 & 39.3 & 32.1 & 9.0 & 86.8 & 99.9 & 17.0 \\ 
 
0.03 & strong2 & 0.1 & 0.01 & 47.6 & 82.2 & 56.8 & 0.3 & 0.0 & 37.6 & 0.0 & 0.0 & 86.8 & 99.9 & 17.0 \\ 
 
0.1 & light0 & 0.0 & 0.03 & 47.6 & 81.9 & 54.4 & 0.5 & 0.0 & 45.9 & 83.1 & 17.0 & 50.7 & 78.7 & 16.0 \\ 
 
0.1 & light0 & 0.0 & 0.01 & 47.6 & 82.9 & 49.5 & 0.1 & 0.0 & 40.9 & 0.0 & 0.0 & 50.7 & 78.7 & 16.0 \\ 
 
0.1 & none & 0.1 & 0.03 & 47.3 & 81.5 & 51.3 & 0.2 & 0.0 & 39.2 & 56.3 & 13.0 & 42.6 & 46.3 & 11.0 \\ 
 
0.1 & none & 0.1 & 0.01 & 47.3 & 82.5 & 47.7 & 0.3 & 0.0 & 33.8 & 0.0 & 0.0 & 42.6 & 46.3 & 11.0 \\ 
 
0.1 & strong1 & 0.1 & 0.03 & 47.1 & 82.3 & 52.6 & 0.1 & 0.0 & 41.0 & 44.6 & 13.0 & 64.2 & 81.9 & 15.0 \\ 
 
0.1 & strong1 & 0.1 & 0.01 & 47.1 & 82.4 & 56.1 & 0.2 & 0.0 & 36.0 & 0.2 & 0.0 & 64.2 & 81.9 & 15.0 \\ 
 
0.1 & strong2 & 0.1 & 0.03 & 46.9 & 82.5 & 51.5 & 0.1 & 0.0 & 41.3 & 47.1 & 15.0 & 67.3 & 92.5 & 17.0 \\ 
 
0.1 & strong2 & 0.1 & 0.01 & 46.9 & 82.4 & 54.6 & 0.2 & 0.0 & 36.0 & 0.0 & 0.0 & 67.3 & 92.5 & 17.0 \\ 
 
0.03 & light0 & 0.0 & 0.03 & 45.0 & 81.5 & 53.5 & 0.3 & 0.0 & 48.7 & 77.7 & 17.0 & 55.8 & 93.0 & 17.0 \\ 
 
0.03 & light0 & 0.0 & 0.01 & 45.0 & 82.5 & 53.4 & 0.3 & 0.0 & 46.9 & 1.0 & 0.0 & 55.8 & 93.0 & 17.0 \\ 
 
0.1 & none & 0.0 & 0.03 & 42.8 & 78.6 & 58.6 & 1.5 & 0.0 & 47.2 & 59.9 & 14.0 & 49.8 & 57.7 & 13.0 \\ 
 
0.1 & none & 0.0 & 0.01 & 42.8 & 80.3 & 53.3 & 2.3 & 1.0 & 41.9 & 0.1 & 0.0 & 49.8 & 57.7 & 13.0 \\ 
 
0.03 & none & 0.0 & 0.03 & 40.1 & 78.1 & 61.7 & 1.2 & 0.0 & 52.3 & 57.4 & 13.0 & 55.8 & 72.0 & 14.0 \\ 
 
0.03 & none & 0.0 & 0.01 & 40.1 & 79.4 & 60.0 & 1.2 & 0.0 & 51.1 & 0.0 & 0.0 & 55.8 & 72.0 & 14.0 

\end{tabular}
}
    \label{tab:big-table-vits}
\end{table}

\begin{table}[]
\setlength{\tabcolsep}{3pt}
    \centering
        \caption{Detailed hyperparameters and AUC values for the ViT-B/16 models with ImageNet-21k pretraining. In the \textit{Mahalanobis (pre)} columns we evaluate the pretraining checkpoint (i.e. \textit{before} finetuning) with Mahalanobis distance. As suggested by \cite{bitterwolf2023ninco}, we also report the number of failed unit tests (with FPR values $\geq10\%$).}
    \rowcolors{2}{white}{lightgray}
\resizebox{1.\textwidth}{!}{%
\begin{tabular}{ccccccccccccccc}

 &  &  &  &  &  & \multicolumn{3}{c}{\textbf{MaxLogit}} & \multicolumn{3}{c}{\textbf{Mahalanobis}} & \multicolumn{3}{c}{\textbf{Maha (pre)}} \\ 
 
    &  &  &  &  &  & NINCO & \multicolumn{2}{c}{unit tests} & NINCO & \multicolumn{2}{c}{unit tests} & NINCO & \multicolumn{2}{c}{unit tests} \\
    
wd & aug & do & adapt-lr & pre-acc & ft-acc & \multicolumn{1}{c}{{AUC [\%]}} & \multicolumn{1}{c}{{AUC [\%]}} & n failed & \multicolumn{1}{c}{{AUC [\%]}}& \multicolumn{1}{c}{{AUC [\%]}} & n failed & \multicolumn{1}{c}{{AUC [\%]}} & \multicolumn{1}{c}{{AUC [\%]}} & n failed  \\ 
  \hline 
0.1 & medium1 & 0.0 & 0.03 & 50.6 & 83.9 & 90.4 & 99.9 & 0.0 & 93.4 & 80.7 & 14.0 & 91.3 & 50.0 & 14.0 \\ 
 
0.1 & medium1 & 0.0 & 0.01 & 50.6 & 84.5 & 90.6 & 99.9 & 0.0 & 94.0 & 99.9 & 0.0 & 91.3 & 50.0 & 14.0 \\ 
 
0.1 & light1 & 0.0 & 0.03 & 50.4 & 83.5 & 90.5 & 99.9 & 0.0 & 93.6 & 77.4 & 16.0 & 92.4 & 53.0 & 16.0 \\ 
 
0.1 & light1 & 0.0 & 0.01 & 50.4 & 84.4 & 91.0 & 99.9 & 0.0 & 94.3 & 99.9 & 0.0 & 92.4 & 53.0 & 16.0 \\ 
 
0.1 & medium2 & 0.0 & 0.03 & 50.4 & 83.9 & 90.4 & 99.9 & 0.0 & 93.4 & 86.3 & 14.0 & 89.7 & 46.1 & 17.0 \\ 
 
0.1 & medium2 & 0.0 & 0.01 & 50.4 & 84.4 & 90.3 & 99.9 & 0.0 & 94.1 & 99.9 & 0.0 & 89.7 & 46.1 & 17.0 \\ 
 
0.03 & medium2 & 0.0 & 0.03 & 50.4 & 84.5 & 91.1 & 99.8 & 0.0 & 93.5 & 85.1 & 13.0 & 79.4 & 33.2 & 17.0 \\ 
 
0.03 & medium2 & 0.0 & 0.01 & 50.4 & 84.0 & 89.4 & 99.8 & 0.0 & 93.4 & 89.8 & 12.0 & 79.4 & 33.2 & 17.0 \\ 
 
0.03 & strong1 & 0.0 & 0.03 & 50.3 & 84.2 & 91.0 & 99.9 & 0.0 & 93.8 & 83.4 & 14.0 & 81.3 & 36.6 & 17.0 \\ 
 
0.03 & strong1 & 0.0 & 0.01 & 50.3 & 84.0 & 89.9 & 99.9 & 0.0 & 94.1 & 90.6 & 11.0 & 81.3 & 36.6 & 17.0 \\ 
 
0.03 & strong2 & 0.0 & 0.03 & 50.2 & 84.1 & 91.1 & 99.9 & 0.0 & 93.6 & 89.4 & 11.0 & 79.5 & 37.9 & 17.0 \\ 
 
0.03 & strong2 & 0.0 & 0.01 & 50.2 & 83.7 & 89.3 & 99.9 & 0.0 & 94.0 & 93.8 & 7.0 & 79.5 & 37.9 & 17.0 \\ 
 
0.1 & strong1 & 0.0 & 0.03 & 49.9 & 83.5 & 90.8 & 99.9 & 0.0 & 93.1 & 86.3 & 9.0 & 91.4 & 59.5 & 14.0 \\ 
 
0.1 & strong1 & 0.0 & 0.01 & 49.9 & 84.0 & 90.1 & 99.8 & 0.0 & 94.0 & 99.9 & 0.0 & 91.4 & 59.5 & 14.0 \\ 
 
0.03 & medium1 & 0.0 & 0.03 & 49.7 & 84.0 & 90.9 & 99.9 & 0.0 & 92.6 & 75.2 & 14.0 & 81.4 & 23.8 & 17.0 \\ 
 
0.03 & medium1 & 0.0 & 0.01 & 49.7 & 84.3 & 90.2 & 99.8 & 0.0 & 92.6 & 85.4 & 13.0 & 81.4 & 23.8 & 17.0 \\ 
 
0.1 & strong2 & 0.0 & 0.03 & 49.6 & 83.6 & 90.1 & 99.8 & 0.0 & 93.8 & 92.0 & 10.0 & 90.1 & 47.7 & 17.0 \\ 
 
0.1 & strong2 & 0.0 & 0.01 & 49.6 & 83.9 & 90.0 & 99.7 & 0.0 & 94.6 & 99.8 & 0.0 & 90.1 & 47.7 & 17.0 \\ 
 
0.03 & light1 & 0.1 & 0.03 & 49.5 & 83.7 & 91.1 & 99.7 & 0.0 & 93.2 & 76.4 & 15.0 & 77.9 & 45.1 & 17.0 \\ 
 
0.03 & light1 & 0.1 & 0.01 & 49.5 & 83.7 & 90.9 & 99.8 & 0.0 & 93.6 & 94.2 & 6.0 & 77.9 & 45.1 & 17.0 \\ 
 
0.1 & light1 & 0.1 & 0.03 & 49.3 & 83.6 & 91.3 & 99.9 & 0.0 & 93.4 & 92.0 & 12.0 & 89.7 & 71.0 & 17.0 \\ 
 
0.1 & light1 & 0.1 & 0.01 & 49.3 & 84.1 & 91.1 & 99.9 & 0.0 & 93.9 & 99.9 & 0.0 & 89.7 & 71.0 & 17.0 \\ 
 
0.03 & medium1 & 0.1 & 0.03 & 49.2 & 83.9 & 90.9 & 99.9 & 0.0 & 93.7 & 94.8 & 9.0 & 80.5 & 44.6 & 17.0 \\ 
 
0.03 & medium1 & 0.1 & 0.01 & 49.2 & 83.4 & 89.9 & 99.9 & 0.0 & 93.7 & 99.6 & 0.0 & 80.5 & 44.6 & 17.0 \\ 
 
0.1 & light0 & 0.1 & 0.03 & 49.0 & 82.7 & 89.7 & 99.9 & 0.0 & 90.8 & 76.1 & 17.0 & 89.3 & 92.3 & 16.0 \\ 
 
0.1 & light0 & 0.1 & 0.01 & 49.0 & 83.3 & 90.5 & 99.8 & 0.0 & 93.1 & 99.3 & 1.0 & 89.3 & 92.3 & 16.0 \\ 
 
0.03 & light0 & 0.1 & 0.03 & 48.9 & 82.9 & 89.9 & 99.9 & 0.0 & 91.2 & 69.6 & 16.0 & 82.8 & 41.9 & 17.0 \\ 
 
0.03 & light0 & 0.1 & 0.01 & 48.9 & 83.2 & 89.9 & 99.9 & 0.0 & 92.2 & 88.6 & 12.0 & 82.8 & 41.9 & 17.0 \\ 
 
0.03 & light1 & 0.0 & 0.03 & 48.5 & 83.7 & 91.5 & 99.7 & 0.0 & 93.0 & 65.6 & 15.0 & 83.1 & 23.0 & 17.0 \\ 
 
0.03 & light1 & 0.0 & 0.01 & 48.5 & 84.2 & 90.9 & 99.8 & 0.0 & 93.5 & 81.4 & 13.0 & 83.1 & 23.0 & 17.0 \\ 
 
0.03 & medium2 & 0.1 & 0.03 & 48.5 & 83.8 & 90.5 & 99.9 & 0.0 & 93.6 & 93.5 & 10.0 & 77.2 & 44.2 & 17.0 \\ 
 
0.03 & medium2 & 0.1 & 0.01 & 48.5 & 83.2 & 89.8 & 99.8 & 0.0 & 94.2 & 99.2 & 0.0 & 77.2 & 44.2 & 17.0 \\ 
 
0.1 & medium1 & 0.1 & 0.03 & 48.2 & 83.5 & 90.5 & 99.9 & 0.0 & 93.1 & 93.4 & 10.0 & 87.1 & 61.7 & 15.0 \\ 
 
0.1 & medium1 & 0.1 & 0.01 & 48.2 & 83.6 & 90.4 & 99.8 & 0.0 & 93.8 & 99.9 & 0.0 & 87.1 & 61.7 & 15.0 \\ 
 
0.03 & strong1 & 0.1 & 0.03 & 48.1 & 83.3 & 90.4 & 99.9 & 0.0 & 92.4 & 85.3 & 15.0 & 77.1 & 45.9 & 17.0 \\ 
 
0.03 & strong1 & 0.1 & 0.01 & 48.1 & 82.8 & 89.8 & 99.7 & 0.0 & 93.1 & 97.7 & 4.0 & 77.1 & 45.9 & 17.0 \\ 
 
0.1 & medium2 & 0.1 & 0.03 & 47.7 & 83.3 & 90.3 & 99.9 & 0.0 & 92.7 & 96.1 & 6.0 & 84.6 & 48.9 & 17.0 \\ 
 
0.1 & medium2 & 0.1 & 0.01 & 47.7 & 83.3 & 89.9 & 99.8 & 0.0 & 93.4 & 100.0 & 0.0 & 84.6 & 48.9 & 17.0 \\ 
 
0.03 & strong2 & 0.1 & 0.03 & 47.6 & 82.9 & 89.7 & 99.9 & 0.0 & 92.4 & 93.3 & 9.0 & 74.1 & 40.4 & 17.0 \\ 
 
0.03 & strong2 & 0.1 & 0.01 & 47.6 & 82.2 & 87.9 & 99.8 & 0.0 & 92.5 & 99.6 & 0.0 & 74.1 & 40.4 & 17.0 \\ 
 
0.1 & light0 & 0.0 & 0.03 & 47.6 & 81.9 & 89.1 & 99.8 & 0.0 & 91.0 & 50.5 & 17.0 & 89.7 & 70.4 & 16.0 \\ 
 
0.1 & light0 & 0.0 & 0.01 & 47.6 & 82.9 & 90.3 & 99.8 & 0.0 & 92.3 & 100.0 & 0.0 & 89.7 & 70.4 & 16.0 \\ 
 
0.1 & none & 0.1 & 0.03 & 47.3 & 81.5 & 89.5 & 99.8 & 0.0 & 92.1 & 77.0 & 13.0 & 91.3 & 88.8 & 11.0 \\ 
 
0.1 & none & 0.1 & 0.01 & 47.3 & 82.5 & 90.3 & 99.8 & 0.0 & 93.5 & 99.9 & 0.0 & 91.3 & 88.8 & 11.0 \\ 
 
0.1 & strong1 & 0.1 & 0.03 & 47.1 & 82.3 & 89.4 & 99.8 & 0.0 & 91.9 & 92.8 & 13.0 & 84.5 & 53.8 & 15.0 \\ 
 
0.1 & strong1 & 0.1 & 0.01 & 47.1 & 82.4 & 88.8 & 99.8 & 0.0 & 92.9 & 99.4 & 0.0 & 84.5 & 53.8 & 15.0 \\ 
 
0.1 & strong2 & 0.1 & 0.03 & 46.9 & 82.5 & 89.6 & 99.8 & 0.0 & 92.1 & 94.8 & 15.0 & 82.9 & 46.1 & 17.0 \\ 
 
0.1 & strong2 & 0.1 & 0.01 & 46.9 & 82.4 & 89.3 & 99.8 & 0.0 & 93.1 & 99.6 & 0.0 & 82.9 & 46.1 & 17.0 \\ 
 
0.03 & light0 & 0.0 & 0.03 & 45.0 & 81.5 & 88.5 & 99.8 & 0.0 & 89.4 & 46.6 & 17.0 & 84.0 & 17.2 & 17.0 \\ 
 
0.03 & light0 & 0.0 & 0.01 & 45.0 & 82.5 & 89.0 & 99.7 & 0.0 & 90.3 & 99.8 & 0.0 & 84.0 & 17.2 & 17.0 \\ 
 
0.1 & none & 0.0 & 0.03 & 42.8 & 78.6 & 87.5 & 99.4 & 0.0 & 89.3 & 55.1 & 14.0 & 88.5 & 57.9 & 13.0 \\ 
 
0.1 & none & 0.0 & 0.01 & 42.8 & 80.3 & 89.0 & 99.4 & 1.0 & 91.7 & 99.9 & 0.0 & 88.5 & 57.9 & 13.0 \\ 
 
0.03 & none & 0.0 & 0.03 & 40.1 & 78.1 & 87.1 & 99.6 & 0.0 & 88.1 & 59.0 & 13.0 & 84.2 & 34.2 & 14.0 \\ 
 
0.03 & none & 0.0 & 0.01 & 40.1 & 79.4 & 87.8 & 99.6 & 0.0 & 89.3 & 99.9 & 0.0 & 84.2 & 34.2 & 14.0 

\end{tabular}
}
    \label{tab:big-table-vits-auroc}
\end{table}

\subsection{Clip Pretraining}
\begin{figure}[htb]
    \centering
    \includegraphics[width=1.\textwidth]{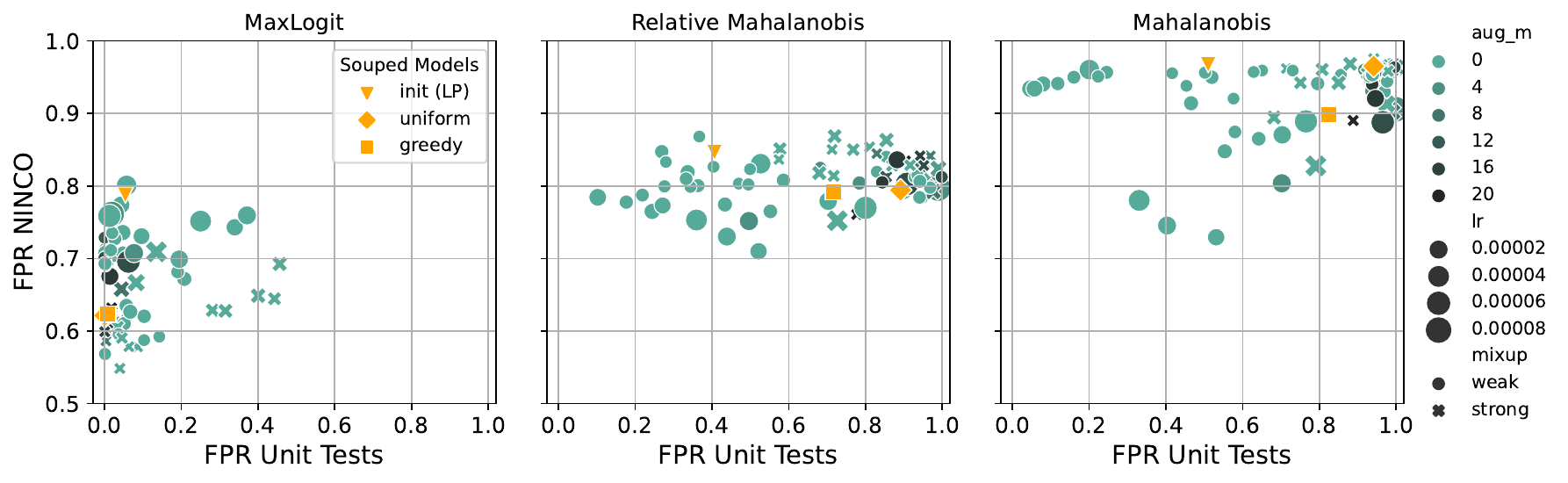}    
    \caption{Clip Models. Souped models are shown in orange.}
    \label{fig:clip}
\end{figure}

In Figure~\ref{fig:clip} we show the plots corresponding to the discussion of the CLIP models in the main paper. 

\FloatBarrier
\subsection{Evaluating more methods}\label{app:more-methods}
\begin{figure}[htb]
    \centering
    \includegraphics[width=1.\textwidth]{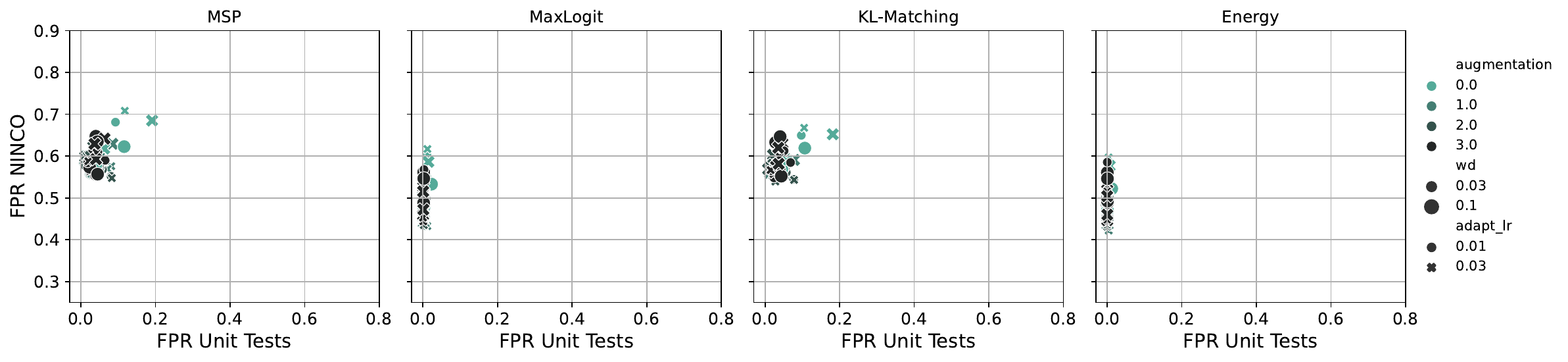}   
    \includegraphics[width=1.\textwidth]{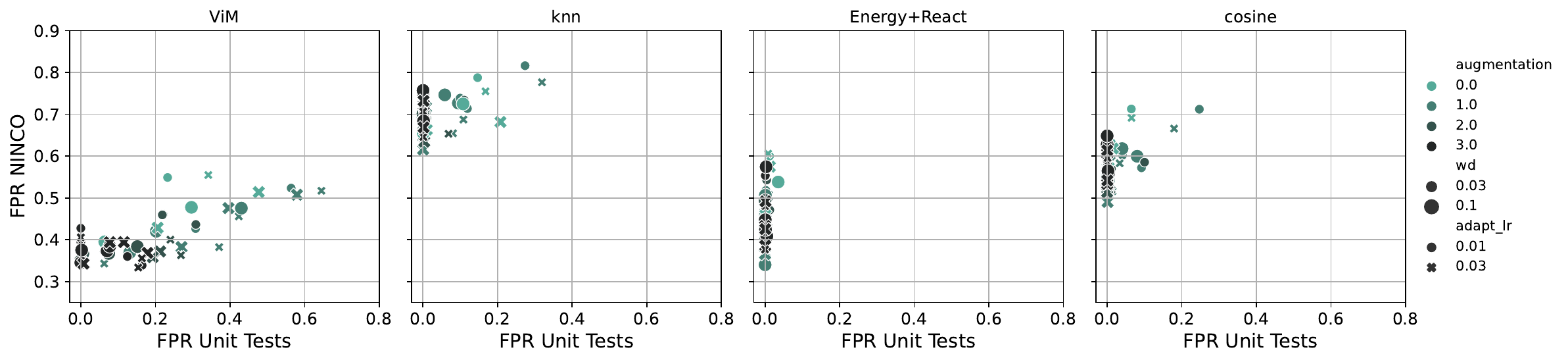}   
    \caption{ImageNet-1k models pretrained on ImageNet-21k with more methods (logit-based on the top, feature-based on the bottom).}
    \label{fig:i21k-more-methods}
\end{figure}

\begin{figure}[htb]
    \centering
    \includegraphics[width=1.\textwidth]{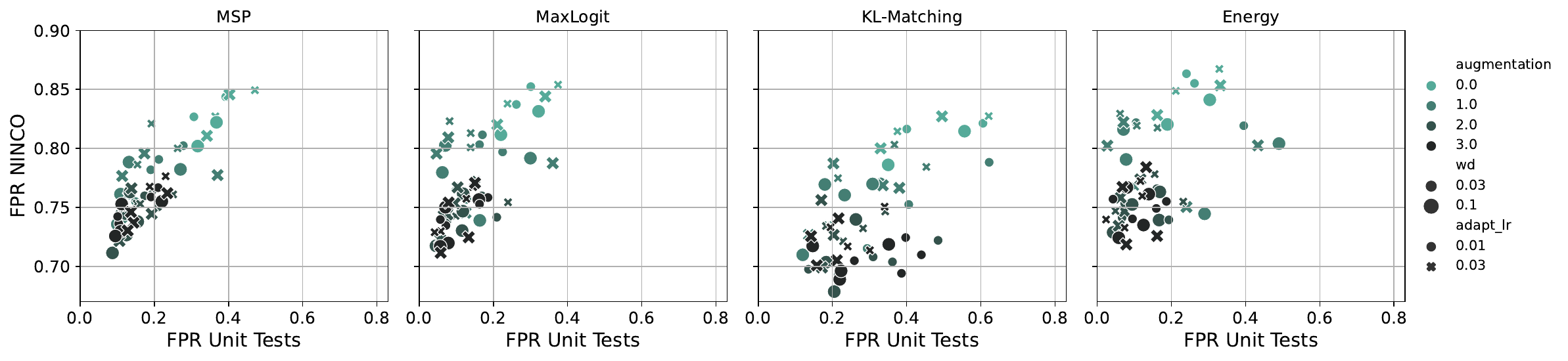}   
    \includegraphics[width=1.\textwidth]{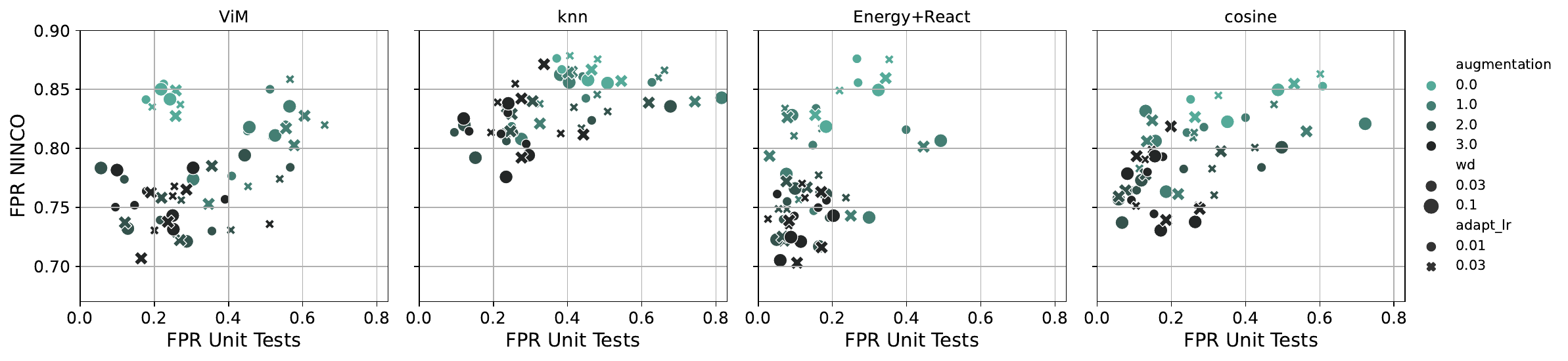}   
    \caption{ImageNet-1k models without pretraining, evaluated with more methods (logit-based on the top, feature-based on the bottom).}
    \label{fig:i1k-more-methods}
\end{figure}

\begin{figure}[htb]
    \centering
    \includegraphics[width=1.\textwidth]{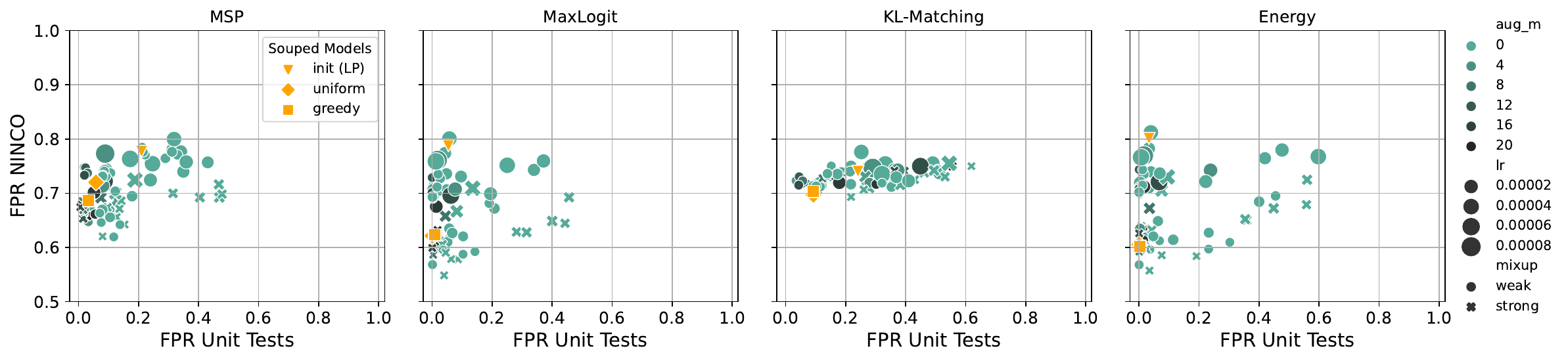}   
    \includegraphics[width=1.\textwidth]{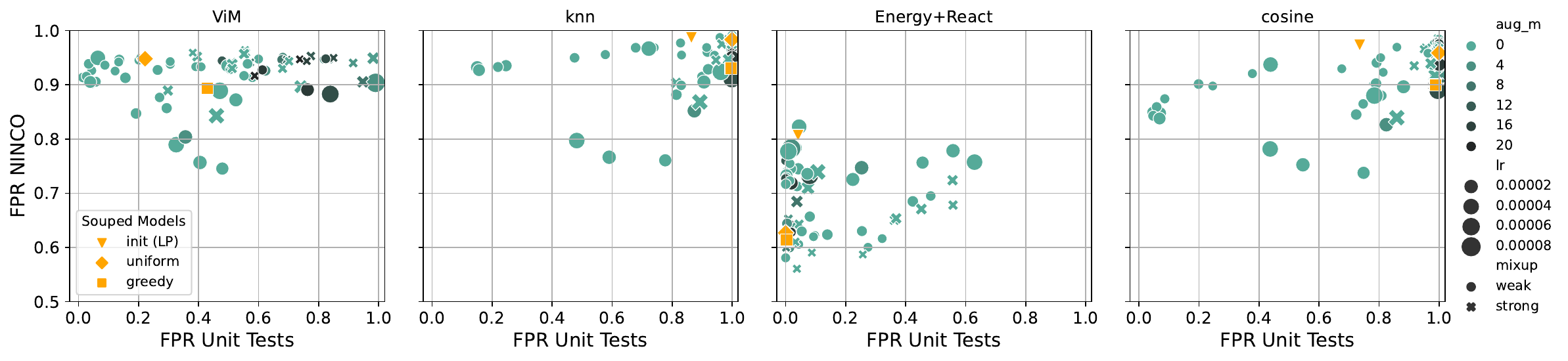}   
    \caption{ImageNet-1k models with CLIP-pretraining evaluated with more methods (logit-based on the top, feature-based on the bottom, souped models in orange).}
    \label{fig:clip-more-methods}
\end{figure}

In Figures~\ref{fig:i21k-more-methods}, \ref{fig:i1k-more-methods}and \ref{fig:clip-more-methods} we plot results for more methods than the ones investigated in the main paper and sketch some of the high-level trends.

\textbf{Pretraining on ImageNet-21k and finetuning on ImageNet-1k .}
The logit-based methods (first row in Figure ~\ref{fig:i21k-more-methods}) all perform relatively well on the unit tests, with MaxLogit and Energy solving the task consistently for all models (FPR values of basically 0\%). For MSP and KL-Matching, the FPR values are not exactly 0, but below 10\% with few exceptions. The results on NINCO vary and do not show conclusive trends with respect to training parameters.

For feature-based methods (second row in Figure~\ref{fig:i21k-more-methods}), the strongest method on NINCO is ViM, where a subset of models achieves FPR values below 40\%, and of those several simultaneously yield close to 0\% FPR on the unit tests. Interestingly, augmentations in combination with dropout now seem to play a more important role than learning rate and weight decay. This is likely due to the fact that ViM uses \textit{both} feature and logit information. Still, many models yield FPR values larger than 20\% FPR.
For Energy+React, consistently good results are achieved on the unit tests, and several models also perform strongly on NINCO (e.g. those trained with augmentation type \textit{light1} and no dropout all achieve FPR values of 34-37\%).  

\textbf{Training exclusively on ImageNet-1k. }
For this pretraining setup (Figure~\ref{fig:i1k-more-methods}), the models do not generally achieve 0\% FPR on the unit tests for any of the displayed methods. In contrast to ImageNet-21k pretraining, feature-based methods loose their edge over logit-based methods on NINCO . For almost all setups, stronger augmentations improve OOD detection.

\textbf{Pretraining on CLIP and finetuning on ImageNet-1k. }
For CLIP-pretrained models (Figure~\ref{fig:clip-more-methods}) out of the feature-based methods only Energy+React works reasonably well on NINCO and the unit tests. For this setup, the souped models (orange in plots) are amongst the better detectors, but they fail completely for e.g. ViM, knn and cosine. For Energy+React and MaxLogit, the overall best detectors, smaller learning rate leads to improved performnace on NINCO.

\FloatBarrier
\section{Methods}\label{sec:methods}
Here we give an overview over the evaluated OOD detection methods.
For clarity, %
we denote vectors in bold and lowercase letters and matrices in bold an uppercase letters. We write neural networks as functions $n$, which are parametrized by weights $\mathbf{\theta}$, take an input sample $\mathbf{x}$ and produce an output vector $\mathbf{o}$ of size $C$, where $C$ is typically the number of classes in a classification task (1000 in the case of IN-1K). We refer to $\mathbf{o}$ as the logits of $\mathbf{x}$, which can be transformed to a probability vector $\mathbf{p}$ (also of size $C$) via the softmax function: $p_{i}=\exp{(o_{i})/\sum_{c}\exp(o_{c})}$. The network $n$ can be decomposed into a feature extractor $h$ and the networks last layer $g$: \begin{equation*}
\mathbf{o}=n(\mathbf{x})=g(h (\mathbf{x})),
\end{equation*}
where $g$ is a fully connected, linear layer, i.e. $g(\mathbf{h})=\mathbf{W}^{T}\mathbf{h}+\mathbf{b}$ with weight $\mathbf{W}$ and bias $\mathbf{b}$. We refer to $\mathbf{h}=h(\mathbf{x})$ as the \textit{features} or the \textit{embeddings} of $\mathbf{x}$ w.r.t. the network $n$.
For each sample $\mathbf{x}$, a method returns an OOD-score $s=f(\mathbf{x})$, a scalar value which is supposed to be larger for ID data and smaller for OOD data.
Methods accessing $h(\mathbf{x})$ directly in order to compute the OOD-score are referred to as feature-based methods, in contrast to methods that derive their OOD-score from the logits $\mathbf{o}$ (even though obviously the logits implicitly also depend on these features). In the following, we will describe how each method computes the score $s$ for a test input $\mathbf{x}$.

\textbf{MSP} \cite{hendrycks2017MSP}: The most popular OOD-detection baseline uses the confidence, i.e. the max softmax probability of a models probability output vector:
\begin{equation*}
s=\max_{c}({p}_{c})
\end{equation*}
\newline
\textbf{Max-Logit} \cite{hendrycks22Scaling}: Similar to MSP, Max-Logit returns the largest entry of the logit-vector $\mathbf{o}$, i.e.  
\begin{equation*}
s=\max_{c}({o}_{c})
\end{equation*}\newline
\textbf{Energy} \cite{liu2020energy}:
The Energy based OOD detection method uses the denominator of the softmax-function as OOD-score:
\begin{equation*}
s=\log\sum_{c}^{C}\exp{(o_{c})}
\end{equation*}
\textbf{KL-Matching} \cite{hendrycks22Scaling}: KL-Matching computes a mean probability vector $\mathbf{d}_c$ for each of the $C$ classes. For a test input, the KL-distances of all $\mathbf{d}_c$ vectors to its probability vector $\mathbf{p}$ are computed, and the OOD-score is the negative of the smallest of those distances:
\begin{equation*}
s=-\min_{c}\text{KL}[\mathbf{p}||\mathbf{d}_{c}]
\end{equation*}
In the original paper by \cite{hendrycks22Scaling}, the average for $\mathbf{d}_{c}$ is computed over an additional validation set. Since none of the other methods leverages extra data and we are interested in fair comparison, we deploy KL-Matching like in \cite{wang2022vim,yang2022openood}, where the average is computed over the train set.

\textbf{KNN} \cite{sun2022knnood}: KNN is a non-parametric method that computes distances in the feature-space. Specifically, the feature vector of a test input is normalized to $\mathbf{z}=\mathbf{h}/||\mathbf{h}||_{2}$ and the pairwise distances $r_{i}(\mathbf{z})=||\mathbf{z}-\mathbf{z}_{i}||_{2}$ to the normalized features $\mathcal{Z}=\{\mathbf{z}_{1},...,\mathbf{z}_{N}\}$ of all samples of the training set are computed.
The distances $r_{i}(\mathbf{z})$ are then sorted according to their magnitude and the $K^{\text{th}}$ smallest distance, denoted $r^{K}(\mathbf{z})$ is used as negative OOD-score:
\begin{equation*}
s=-r^{K}(\mathbf{z})
\end{equation*}
Like suggested in \cite{sun2022knnood}, we use $K=1000$.\newline
\textbf{Mahalanobis distance} \cite{LeeMahalanobis2018}: This popular method fits a class-conditional Gaussian with shared covariance matrix to the train set, i.e. computes
\begin{equation*}
\hat{\mu}_{c}=\frac{1}{N_{c}}\sum_{i:y_{i}=c}\mathbf{h}_{i},\hspace{15pt} \hat{\mathbf{\Sigma}}=\frac{1}{N}\sum_{c}\sum_{i:y_{i}=c}(\mathbf{h}_{i}-\hat{\mu}_{c})(\mathbf{h}_{i}-\hat{\mu}_{c})^{T}
\end{equation*}
where $N_{c}$ is the number of train samples in class $c$ and $N$ is the total number of train samples. The OOD-score of a test sample is then the Mahalanobis distance induced by $\hat{\mathbf{\Sigma}}$ between its feature $\mathbf{h}$ and the closest class mean:
\begin{equation*}
s=-\min_{c}(\mathbf{h}-\hat{\mu}_{c})\hat{\mathbf{\Sigma}}^{-1}(\mathbf{h}-\hat{\mu}_{c})^{T}
\end{equation*}

\textbf{Relative Mahalanobis distance} \cite{RenRelMaha2021}: A modification of the Mahalanobis distance method, thought to improve near-OOD detection, is to additionally fit a global Gaussian distribution to the train set without taking class-information into account: 
\begin{equation*}
\hat{\mu}_{\text{global}}=\frac{1}{N}\sum_{i}\mathbf{h}_{i},\hspace{15pt} \hat{\mathbf{\Sigma}}_{\text{global}}=\frac{1}{N}\sum_{i}(\mathbf{h}_{i}-\hat{\mu}_{\text{global}})(\mathbf{h}_{i}-\hat{\mu}_{\text{global}})^{T}
\end{equation*}
The OOD-score is then defined as the difference between the original Mahalanobis distance and the Mahalanobis distance w.r.t. the global Gaussian distribution:
\begin{equation*}
s=-\min_{c}\left( (\mathbf{h}-\hat{\mu}_{c})\hat{\mathbf{\Sigma}}^{-1}(\mathbf{h}-\hat{\mu}_{c})^{T}-(\mathbf{h}-\hat{\mu}_{\text{global}})\hat{\mathbf{\Sigma}}_{\text{global}}^{-1}(\mathbf{h}-\hat{\mu}_{\text{global}})^{T}\right)
\end{equation*}

\textbf{ReAct} \cite{sun2021react}: The authors propose to perform a truncation of the feature vector, $\mathbf{\Bar{h}}=\min(\mathbf{h},r)$, where the $\min$ operation is to be understood element-wise and $r$ is the truncation threshold.  The truncated features can then be converted to so-called rectified logits via $\mathbf{\Bar{o}}=g(\mathbf{\Bar{h}})=\mathbf{W}^{T}\mathbf{\Bar{h}}+\mathbf{b}$. While the rectified logits can now be used with a variety of existing detection methods, we follow \cite{sun2021react} and use the rectified Energy as OOD-score:
\begin{equation*}
s=\log\sum_{c}^{C}\exp{(\Bar{o}_{c})}
\end{equation*}
As suggested in \cite{wang2022vim}, we set the threshold $r$ such that 1\% of the activations from the train set would be truncated.\newline

\textbf{Virtual Logit Matching} \cite{wang2022vim}: The idea behind ViM is that meaningful features are thought to lie in a low-dimensional manifold, called the principal space $P$, whereas features from OOD-samples should also lie in $P^{\perp}$, the space orthogonal to $P$. $P$ is the $D$-dimensional subspace spanned by the eigenvectors with the largest $D$ eigenvalues of the matrix $\mathbf{F}^{T}\mathbf{F}$, where $\mathbf{F}$ is the matrix of all train features offsetted by $\mathbf{u}=-(\mathbf{W^{T}})^{+}\mathbf{b}$ ($+$ denotes the Moore-Penrose inverse). A sample with feature vector $\mathbf{h}$ is then also offset to $\mathbf{\Tilde{h}}=\mathbf{h}-\mathbf{u}$ and can be decomposed into $\mathbf{\Tilde{h}}=\mathbf{\Tilde{h}}^{P}+\mathbf{\Tilde{h}}^{P^{\perp}}$, and $\mathbf{\Tilde{h}}^{P^{\perp}}$ is referred to as the \textit{Residual} of $\mathbf{h}$. ViM leverages the Residual and converts it to a virtual logit $o_{0}=\alpha||\mathbf{\Tilde{h}}^{P^{\perp}}||_{2}$, where 
\begin{equation*}
\alpha=\frac{\sum_{i=1}^{N}\max_{c}o_{i}^{c}}{\sum_{i=1}^{N}||\mathbf{h}^{P^{\perp}}_{i}||_{2}}
\end{equation*}
is designed to match the scale of the virtual logit to the scale of the real train logits. The virtual logit is then appended to the original logits of the test sample, i.e. to $\mathbf{o}$, and a new probability vector is computed via the softmax function. The probability corresponding to the virtual logit is then the final OOD-score:
\begin{equation*}\label{vim}
s=-\frac{\exp{(o_0)}}{\sum_{c=1}^{C}\exp{(o_{c})}+\exp{(o_0)}}
\end{equation*}
Like suggested in \cite{wang2022vim}, we use $D=1000$ if the dimensionality of the feature space $d$ is $d\geq2048$, $D=512$ if $2048\geq d\geq 768$, and $D=d/2$ rounded to integers otherwise. 

\textbf{Cosine} \cite{tack2020csi,anonymous2023COOD}: This method computes the maximum cosine-similarity between the features of a test-sample and embedding vectors $\Tilde{\mathbf{u}}_{c}$ (sometimes also called concept-vector):
\begin{equation}
    s=\max_{c}\Tilde{\mathbf{u}}_{c}^{T}\mathbf{h}/||\Tilde{\mathbf{u}}_{c}^{T}||_{2}
\end{equation}
For zero-shot CLIP, $\Tilde{\mathbf{u}}_{c}$ can be obtained by creating text-embeddings from the ImageNet class names. Encoding \textit{'A photo of a ...'} yields an embedding from the corresponding class. For classifiers, we use the class-wise train means $\hat{\mu}_c$, that are also used for Mahalanobis distance.

\FloatBarrier

\section{Definitions of OOD detection metrics}\label{sec:fpr_def}
The performance of OOD detectors is commonly reported in terms of the \textit{false positive rate at a fixed true positive rate} Q, denoted as \textbf{\FPRQ}, short \textbf{\FPR}.
This means that the detector is interpreted as %
making the decision to \textit{accept} an unknown input $x$ \textit{if} $S(x) \geq \tau$, for a threshold $\tau$ that is chosen such that Q\% of ID inputs are accepted, and \textit{rejecting} the input as OOD \textit{if} $S(x) < \tau$.
The \FPRQ{} counts the fraction of falsely accepted OOD inputs under this decision scheme.
This means the \textit{lower} the \FPRQ{}, the \textit{better} the OOD detection performance.
In the OOD detection literature, the most commonly used value for Q is 95\%, which we too use throughout this paper. As suggested in \cite{bitterwolf2023ninco}, we compute the FPR value for each NINCO class separately, and report the average FPR across classes. For the unit-tests, the authors suggest reporting the number of failed unit tests (defined by a FPR threshold of $10\%$). Since this would lead to overlapping data points in our plots, we decided to show the mean FPR across the unit-test classes instead in the figures. For Table~\ref{tab:big-table-vits}, we also report the number of failed unit tests.  

\end{document}